\documentclass[runningheads]{llncs}
\usepackage[T1]{fontenc}
\usepackage{graphicx}
\usepackage{booktabs}
\usepackage[misc]{ifsym}
\usepackage{arydshln}

\usepackage{mwe}

\newcommand{\bs}{\boldsymbol}
\newcommand{\loss}{\mathcal{L}} 
\newcommand{\setr}{\mathcal{R}} 
\newcommand{\setp}{\mathcal{P}}
\newcommand{\tensor}[1]{\bs{\mathcal{#1}}} 
\newcommand{\setw}[1]{\widehat{\bs{\mathcal{#1}}}}

\newcommand{\approach}{\texttt{RENE}}

\usepackage[linesnumbered,ruled,vlined]{algorithm2e}
\usepackage{amsmath}
\usepackage{amssymb}
\usepackage{xcolor}
\usepackage{amsfonts}
\usepackage{stmaryrd}
\usepackage{wrapfig}
\usepackage{caption}
\usepackage{lipsum}
\usepackage{multirow}
\usepackage{siunitx}
\usepackage{tabularx}
\usepackage{booktabs}
\usepackage{pifont}
\usepackage{adjustbox} 
\usepackage{array}     

\begin{document}

\title{Unified Framework for Pre-trained Neural Network Compression via Decomposition and Optimized Rank Selection}

\titlerunning{Rank adapt tensor decomposition}


\author{Ali Aghababaei-Harandi\and
Massih-Reza Amini}


\institute{Université Grenoble Alpes, CNRS, \\Computer Science Laboratory LIG, Grenoble, France \\ \email{\{Firstname.Lastname}\}@univ-grenoble-alpes.fr}



\maketitle              

\begin{abstract}
Despite their high accuracy, complex neural networks demand significant computational resources, posing challenges for deployment on resource constrained devices such as mobile phones and embedded systems. Compression algorithms have been developed to address these challenges by reducing model size and computational demands while maintaining accuracy. Among these approaches, factorization methods based on tensor decomposition are theoretically sound and effective. However, they face difficulties in selecting the appropriate rank for decomposition. This paper tackles this issue by presenting a unified framework that simultaneously applies decomposition and  rank selection, employing a composite compression loss within defined rank constraints. Our method includes an automatic rank search in a continuous space, efficiently identifying optimal rank configurations for the pre-trained model by eliminating the need for additional training data and reducing computational overhead in the search step. Combined with a subsequent fine-tuning step, our approach maintains the performance of highly compressed models on par with their original counterparts. Using various benchmark datasets and models, we demonstrate the efficacy of our method through a comprehensive analysis.

\keywords{Neural Network, Decomposition, Optimal Rank.}
\end{abstract}

\section{Introduction}
In recent years, deep learning has revolutionized various scientific fields, including computer vision and natural language processing \cite{radford2021learning}. Complex neural networks with millions or billions of parameters have achieved unprecedented accuracy. However, their size poses challenges for deployment on resource-limited devices like mobile phones and edge devices \cite{sandler2018mobilenetv2}. The storage, memory, and processing requirements of these models often prove to be unfeasible or excessively costly, thus limiting their practicality and accessibility.

Recent research has introduced various compression algorithms to address cost-effectiveness, scalability, and real-time responsiveness \cite{novikov2015tensorizing}. These approaches, which reduce a model's size and computational demands while preserving accuracy, can be classified into four primary categories. One straightforward method is \textit{pruning}, which involves removing insignificant weights from the model \cite{blalock2020state}. \textit{Quantization} reduces the precision of numerical values, typically transitioning from 32-bit floating-point numbers to lower bit-width fixed-point numbers \cite{park2017weighted}. \textit{Knowledge distillation} trains a smaller ``student'' model to mimic a larger ``teacher'' model, resulting in a compact model with similar performance \cite{beyer2022knowledge}. Lastly, \textit{low-rank factorization} decomposes weight matrices or tensors into smaller components, reducing the number of parameters \cite{audibertACl23,yang2017tensor,yin2022batude}. While effective, selecting the appropriate rank for decomposition remains a significant challenge.

Non-uniqueness in tensor rank is a major challenge in tensor decomposition research. Most tensor decomposition problems, especially CP decomposition, are NP-hard \cite{hillar2013most}, and allow different decompositions of a same tensor even though some works tries to approximating the ranks of a tensor in a practical way \cite{xu2023tensor,goldfarb2014robust}. Finding the ideal rank is an ongoing research topic, and determining multiple tensor ranks for deep neural network layers is not suitable for conventional hyperparameter selection methods like cross-validation. Typically, a single rank is chosen for the decomposition of layers based on a compression rate, but this can lead to significant performance degradation in complex models.

Recent studies propose automated methods for determining tensor decomposition ranks \cite{ijcai2024p418,li2021heuristic,xiao2023haloc}. However, these approaches, including reinforcement learning, greedy search algorithms, and SuperNet search, can be computationally expensive and time-consuming, especially for large models and datasets. Their effectiveness often depends on hyperparameters like learning rates or regularization parameters, which are challenging to tune. Additionally, existing methods do not cover a wide enough search space to achieve ideal compression rates.

This paper introduces a unified framework that simultaneously addresses tensor decomposition and optimal rank selection using a composite compression loss within specified rank constraints. Also, when we combine this rank search with a subsequent fine-tuning step, our experiments show that the highly compressed model performs similarly to the original model. The key contributions of this paper are:
\begin{itemize}
\item Our proposed method allows to achieve maximum compression rates by covering all ranks in the search space through a simple and efficient multi-step search process that explores ranks from low to high resolution.

\item The proposed search method involves an automatic rank search in a continuous space, which efficiently identifies the optimal rank configurations for layer decomposition without requiring training data.

\item We perform a comprehensive analysis of the various components of our approach, highlighting its efficacy across various benchmark datasets and models such as convolution and transformer-based models. we achieved improvement in some experiments specifically improvement in all metrics in the case of ResNet-18, while in another experiment we had competitive results. Moreover, our method speeds up the search phase compared to other related work.
\end{itemize}

\section{Related Work}
Low-rank factorization techniques, particularly tensor decomposition, have gained attention in deep learning, especially in natural language processing (NLP) \cite{novikov2015tensorizing}. These methods provide an efficient means of fine-tuning large language models, offering advantages over alternative techniques such as quantization \cite{park2017weighted}, knowledge distillation \cite{beyer2022knowledge}, and gradient-based pruning \cite{yin2021towards}. In this paper, we focus on tensor decomposition, which proved to be a robust compression tool with a high compression rate and a relatively lower computational cost. Their applications extend beyond NLP and have also been applied in computer vision \cite{yin2022batude}. However, selecting the appropriate rank for compressing deep neural models using decomposition techniques is NP-hard \cite{hillar2013most}. Research in this area falls into two main approaches.

The first approach relies on a rank-fixed setting, where the ranks of layers are determined based on a predefined compression rate target. Some work used a low-rank loss to substitute the weights of convolution layers with their low-rank approximations \cite{yu2017compressing}. The two main low-rank approximation methods applied on pre-trained models are CP and Tucker decomposition \cite{kim2015compression}. Recent studies have revealed that fine-tuning after CP decomposition can be unstable and have addressed this issue by integrating a stability term into the decomposition process \cite{phan2020stable}. 
In addition, some work decomposed convolution and fully connected layers with tensor train, and trained the model from scratch  \cite{novikov2015tensorizing}. However, tensor decomposition in a fixed-rank setting presents certain challenges. First, selecting the appropriate rank for different layers is complex and often relies on human expertise. Second, there is a lack of interpretable patterns between layer ranks, leading to inconsistencies among the chosen ranks between layers. Furthermore, the fixed rank strategy overlooks the varying importance of layers \cite{filters2016pruning}, which can result in suboptimal approximations that can lead to accuracy drops or insufficient compression rates.

The second approach involves determining the optimal ranks by setting the optimization problem on the basis of the ranks of layers. One technique consists in iteratively decreasing the ranks of the layers at each step of the search phase \cite{gusak2019automated}. The discrete nature of rank search lends itself to discrete search algorithms, such as reinforcement learning and progressive search, to identify optimal ranks \cite{li2021heuristic}. Other methods impose constraints on ranks and budget, using iterative optimization strategies \cite{yin2021towards}. More recent studies explore continuous search spaces to determine optimal ranks \cite{dai2023deep,xiao2023haloc,XiaoYGZR023,chang2024flora}.

To address time complexity issues, these approaches depend on training data to search for ranks, restricting exploration to a limited search space, and thereby limiting the achievable compression rate. In contrast, we introduce a novel optimization problem that minimizes a decomposition loss while enforcing a rank loss constraint independent of the training data, which accelerates the search process for large models. For rank selection, we propose an efficient dichotomous search method that is both fast and allows for a broader range of rank exploration, ultimately enhancing the compression rate.


\section{Background and Preliminaries}
In the following, we represent indices using italicized letters and sets with italic calligraphic letters. For two-dimensional arrays (matrices) and one-dimensional arrays (vectors), we use bold capital letters and bold lowercase letters, respectively. Finally, tensors are represented as multidimensional arrays with bold calligraphic capital letters.

A fundamental technique for efficiently representing and processing tensors is tensor decomposition. This technique transforms a multidimensional array of data into a series of lower-dimensional tensors, thereby reducing both the representation size and computational complexity.  The prevalent tensor decomposition techniques encompass canonical polyadic (CP) \cite{DOMANOV2017342}, Tucker \cite{tucker1966some}, tensor train (TT) \cite{oseledets2011tensor}, and tensor ring (TR) decomposition~\cite{novikov2015tensorizing}. 

In our work, we employ both the TT and CP decompositions. TT decomposition supports fast multilinear multiplication and integration while preserving structure, and CP decomposition has been shown to achieve high parameter reduction in CNNs with small performance drops \cite{novikov2015tensorizing}. In the following, we present the TT decomposition due to its structural advantages in capturing complex dependencies. 

TT decomposition decomposes a tensor into smaller tensors with dimensions connected as a chain to each other. This decomposition mathematically can be represented as follows:
\begin{align}
    \label{eq:TTD}
    \hat{\tensor{W}}^{(R_1,\ldots,R_{N-1})}(i_1,i_2,...,i_N) 
    &=  \sum_{j_1=1}^{R_1}\cdots \sum_{j_{N-1}=1}^{R_{N-1}}  
    \tensor{G}_1(i_1,j_1) \tensor{G}_2(j_1,i_2,j_2) \notag \\
    &\quad \cdots \tensor{G}_N(j_{N-1},i_N),  
\end{align}
where the tuple $(R_1,R_2,\ldots,R_{N-1})$ represents the rank of the TT decomposition, and $\mathcal{G}_k$ are the TT cores with sizes $R_{k-1}\times I_k\times R_k$, and $R_0=R_N=1$. For a given convolutional layer with a weight tensor $\tensor{W} \in {R}^{b\times h\times w\times c}$, the forward process for an input tensor $\tensor{X} \in {R}^{k_1\times k_2\times k_3}$ can be expressed as:

\begin{figure}[t!]
    \centering
        \includegraphics[width=.55\textwidth]{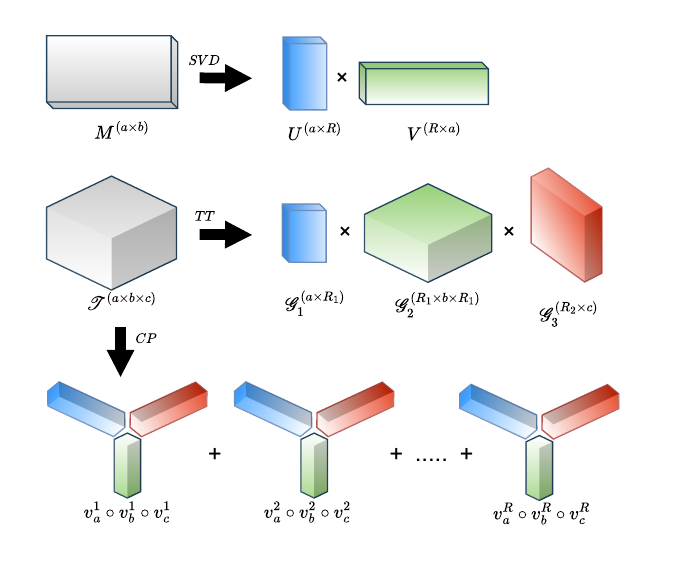}
        \caption{An illustration of matrix decomposition (\textbf{upper row}) using SVD for a matrix $M \in \mathbb{R}^{a \times b}$, alongside
        Tensor Train decomposition (\textbf{middle row}), 
        and CP decomposition (\textbf{bottom row}) 
        for a tensor $\tensor{T} \in \mathbb{R}^{a \times b \times c}$.}
        \label{fig:cp-tt}
\end{figure}
\begin{equation}
\label{forward}
\tensor{Y} =\sum_{i_1=0}^{k_1-1}\sum_{i_2=0}^{k_2-1}\sum_{i_3=0}^{k_3-1} \tensor{W}(t,x+i_1,y+i_2,z+i_3) \tensor{X} (i_1,i_2,i_3).    
\end{equation}
Specifically, we investigate how the weight tensor of a convolutional layer can be decomposed into multiple smaller convolution operations. We utilize TT decomposition, as detailed in the following formulations:
\begin{align}
\label{tt:forward}
\tensor{Y} &= \sum_{r_1=1}^{R_1}\sum_{r_2=1}^{R_2} \tensor{G}_t(t,r_1) 
\bigg(\sum_{i_1=0}^{k_1-1}\sum_{i_2=0}^{k_2-1} \tensor{G}_y(r_1,x+i_{1},y+i_{2},r_2) \notag \\
&\quad \bigg (\sum_{i_{3}=0}^{k_3-1} \tensor{G}_s(i_3,r_2) \tensor{X} (i_1,i_2,i_3) \bigg) \bigg).    
\end{align}
The CP decomposition expresses a multi-dimensional tensor into a sum of rank-one tensors. It follows a well-established factorization process that has been extensively studied in prior works \cite{DOMANOV2017342}. Figure \ref{fig:cp-tt} illustrates TT decomposition and CP decomposition in relation to matrix decomposition using SVD.

\section{Optimal Rank Tensor Decomposition}
\begin{figure}[t!]
    \centering
    \includegraphics[width=.75\textwidth, trim=30 0 0 0, clip]{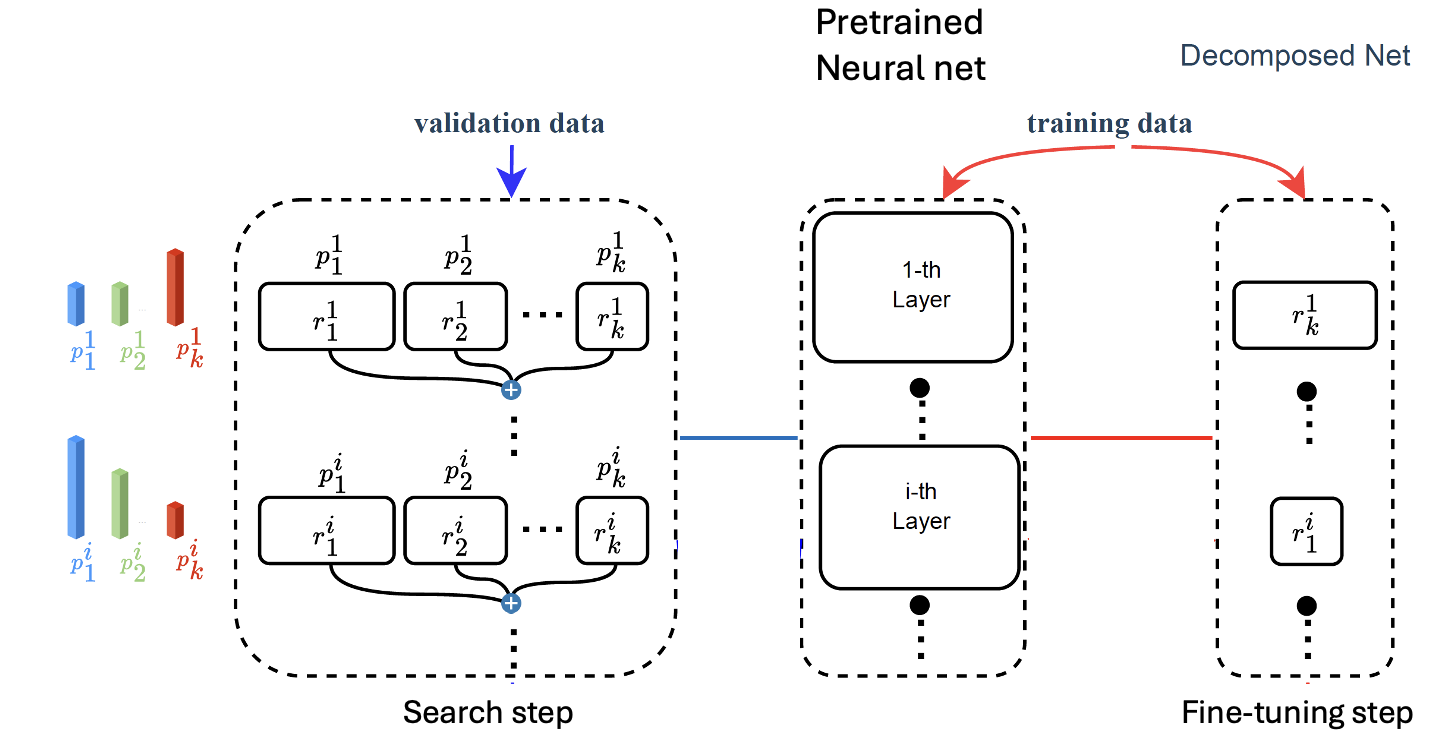} 
    \caption{Overview of \approach{}: Starting with a pre-trained neural network, weight tensors are decomposed layer by layer into lower-rank approximations. Rank coefficients for each layer are refined until optimal (left), followed by fine-tuning of the decomposed network (right).}
    \label{unified}
\end{figure}

The proposed method, denoted as Rank adapt tENsor dEcomposition (\approach) and illustrated in Figure \ref{unified}, involves tensor decomposition with an automatic search for optimal ranks. The approach begins with a pre-trained neural network and aims to decompose its weight tensors layer by layer into lower-rank approximations while minimizing both decomposition and rank losses. This is achieved through an iterative optimization process that updates the decomposition weights and rank coefficients. 

At each layer $i\in\{1,\ldots,n\}$, rank coefficients $(p_j^{i})_j$ related to a set of ranks $\setr_i$ for decomposition (Figure \ref{unified} (left)) are found iteratively and progressively refined until a single optimal rank is determined. The decomposed network with this optimal rank is fine-tuned to align its outputs with the original model (Figure \ref{unified} (right)), ensuring that the compressed model retains the performance of the original while being more efficient.

Equations \eqref{eq:TTD} and \eqref{tt:forward} show that both the number of parameters and computation complexity are directly proportional to the rank of the layer. Consequently, selecting a lower rank results in a reduction in these computational costs. From this observation, we define the decomposition problem as the minimization of a decomposition error under a rank constraint. 

\subsection{Problem Formulation}

Given a pre-trained neural network with $n$ hidden layers and weights $\{\tensor{W}_i\}_{i=1}^{n}$, our objective is then to achieve a low-rank decomposition of these weights with the smallest possible ranks, formulated as the following optimization problem:

\begin{equation}
\label{general}
\underset{\setw{W}^{\setr}}{\min} ~\mathcal{L}_d(\setw{W}^{\setr}) \quad  s.t. \quad \underset{\setr}\min ~\mathcal{L}_{r}(\setr),   
\end{equation}

where $\mathcal{L}_d(.)$ and $\mathcal{L}_r(.)$ are a decomposition loss and a rank loss, respectively, and $\setw{W}^{\setr}=~\left\{\hat{\tensor{W}}^{(\setr_1)}_1,\ldots, \hat{\tensor{W}}^{(\setr_n)}_n \right\}$ is the set of decompositions to be found with $\setr=\{\setr_1,\ldots,\setr_n\}$ the set of ranks, and $\hat{\tensor{W}}^{(\setr_i)}_i$ are the weights of decomposition corresponding to the ranks $\setr_i = \{r_{i}^{1},...,r_{i}^{k}\}$ of layer $i$.

For each layer $i\in\{1,\ldots,n\}$ of the network, we consider a set of decompositions $(\hat{\tensor{W}}_i^{(r)})_{r\in\setr_i}$ of varying ranks defined in the set $\setr_i$, for each weight tensor~$\tensor{W}_i$. To make the optimization problem under the rank constraint \eqref{general} continuous, we associate a rank coefficient $p_i^{(r)}$ with each decomposition of rank $r$ in layer $i$ based on a learnable parameter $\alpha_i^{(r)}$. 

This rank coefficient, defined as $p_i^{(r)}=~\text{softmax}(\alpha_i^{(r)})$, is adjusted via the parameter $\alpha_i^{(r)}$ to reflect the probability that the rank $r$ will be used in the decomposition of the weight tensor $\mathcal{W}_i$ for the layer $i$. Inspired by \cite{wu2019fbnet}, we formalize the rank constraint in \eqref{general} using a normalized rank loss:

\begin{equation}
\label{eq:LossR}
\loss_{r}(\mathcal{R}) =\gamma\sum_{i=1}^n\left(\sum_{r\in \setr_i} p_{i}^{(r)} \frac{r}{\max{\setr_i}}    \right)^\beta,
\end{equation}
where $\beta, \gamma\in[0,1]$ are hyperparameters.

\subsection{Tensor Decomposition and Rank Exploration}

Building on the definition of $\loss_{r}(\mathcal{R})$, we introduce two total losses to update weights and parameters $(\alpha_i^{(r)})_{i,r}$.  The total weights loss for a neural network model with $n$ layers is defined as follows:
\begin{equation}
\label{eq:TotalLoss}
\mathcal{L}_{Tw}(\setw{W}^{\setr}, \setp^{\setr}) = \sum_{i=1}^{n}\!\left\|\bs{\mathcal{W}}_i-\sum_{r\in \setr_i}  p_{i}^{(r)}\hat{\bs{\mathcal{W}}}^{(r)}_{i}\right\|^2_F \!\! \times \gamma\left[\! \sum_{i=1}^{n}\left(\sum_{r\in \setr_i} p_{i}^{(r)} \frac{r}{\max \setr_i}\right)^{\beta}  \right], 
\end{equation}
and the total parameters loss for the same neural network can be formulated as:
\begin{equation}
\label{eq:TotalLossalpha}
\mathcal{L}_{T\alpha}(\setw{W}^{\setr}, \setp^{\setr}) = \mathcal{L}_{val} (\setw{W}^{\setr}, \setp^{\setr})\times \gamma \left[\sum_{i=1}^{n}\left(\sum_{r\in \setr_i} p_{i}^{(r)} \frac{r}{\max\setr_i}\right)^{\beta}  \right],
\end{equation}

The first term in \eqref{eq:TotalLoss} is referred to as the \textit{decomposition loss}, while $\mathcal{L}_{\text{val}}$ in \eqref{eq:TotalLossalpha} denotes the cross-entropy loss on the validation data.
 The minimization of the losses, is tackled through a two-step iterative process. First, the weights of the decomposition, denoted as  $\setw{W}^{\setr}$, are updated by minimizing \eqref{eq:TotalLoss} while keeping the rank coefficients, denoted as $\setp^{\setr}$, fixed across all layers. 
Next, the parameters $(\alpha_i^{(r)})_{i,r}$, are updated using the newly updated weights $\setw{W}^{\setr}$. This update is performed by minimizing the \eqref{eq:TotalLossalpha}, where the weights between the layers are adjusted with the corresponding rank parameters. This two-steps updates ensures that each $\alpha$ update is based on well-trained weights, avoiding noisy signals from still-learning weights. This prevents the architecture from overfitting to transient weight states.

The updates of the decomposition weight parameters and rank coefficients are performed using stochastic gradient descent to ensure efficient and iterative optimization. The update rules are as follows:

\begin{eqnarray}
\text{Weight update:}& ~\hat{\tensor{W}}^{(r)}_{i}&\!\!\gets\!\hat{\tensor{W}}^{(r)}_{i}-\eta_w \nabla_{\hat{\tensor{W}}_i^{(r)}}  (\mathcal{L}_{Tw}), \label{eq:WU}\\
\text{Rank coefficient update:}& ~\alpha^{(r)}_{i}&\!\!\gets\!\alpha^{(r)}_{i}-\eta_{\alpha} \nabla_{\alpha^{(r)}_{i}} (\mathcal{L}_{T\alpha}). \label{eq:PU}
\end{eqnarray}

For each layer $i$ and each rank $r\in \setr_i$, the weight update \eqref{eq:WU} and rank coefficient update \eqref{eq:PU} are performed iteratively until a local minimum of the total loss \eqref{eq:TotalLoss} is reached. Each loss is a multiplication combination, where decomposition and validation losses are scaled by the rank loss (and vice versa). This scale-invariant feature balances both terms without requiring separate trade-off hyperparameters, enabling more stable training and better results compared to additive combinations.

\subsection{Rank Search Space}
Previous approaches to rank search in neural network compression typically rely on evaluating a small, fixed set of candidate ranks. While this strategy offers computational efficiency, it risks overlooking the most optimal rank configurations, as the true optimum may lie between the preselected candidates. To address this limitation, we propose a multi-step rank search method that systematically explores the entire rank space and progressively refines the search around the most promising solutions.

The process begins by defining a broad search space for each layer, denoted as $\mathcal{R}_i$, which spans all feasible rank values from $r_{\min}$ to $r_{\max}$. An initial step size $s^{(0)}$ is chosen to sample candidate ranks at regular intervals across this range, ensuring a coarse but complete coverage of the search space. For each sampled rank, the network weights and associated rank coefficients are updated, allowing the model to adapt to the current rank configuration. The quality of each candidate rank is assessed according to a loss function that may include both reconstruction error and a regularization term to encourage lower ranks.

After this initial exploration, the method identifies, for each layer $i$, the rank $\bar{r}_i$ that achieves the highest rank coefficient, indicating its potential as a promising candidate. To focus the search more precisely, the algorithm then defines new lower and upper bounds, $Lb_i$ and $Ub_i$, centered around $\bar{r}_i$ and separated by half the previous step size on either side. This effectively narrows the search space to a region most likely to contain the optimal rank. The step size is then reduced by a factor $f > 1$, yielding a finer sampling resolution for the next iteration. The new set of candidate ranks for layer $i$ is thus given by:
\[
\mathcal{R}_i=\{r \mid r=Lb_i + ks, \text{ for } k\in\mathbb{N}, \text{ and } Lb_i\leq r\leq Ub_i\},
\]

where $s$ denotes the updated step size. Before commencing the next iteration, the weights and rank coefficients are reinitialized for the refined search space. The process of sampling, updating, and selecting is then repeated. With each iteration, the search space contracts and the step size decreases, leading to an increasingly precise localization of the optimal rank. This iterative refinement continues until, for each layer, the candidate set $\mathcal{R}_i$ contains only a single element, signifying convergence to a unique rank selection.

\begin{figure}
    \centering
    \vspace{-25pt}
    \includegraphics[scale=1]{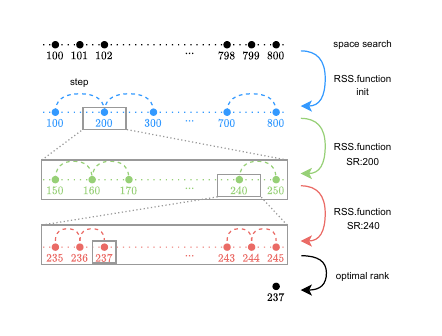}
    \vspace{-20pt}
    \caption{A toy example illustrates the search for rank spaces. Initially, the search space includes integers from  100 to 800, with a step size of 100. After the first iteration, the selected rank is $\bar{r}=200$, narrowing the search interval to $[150,250]$ with a step size of $10$. The second iteration selects $\bar{r}=240$, refining the search space to $[235,245]$ with a step size of $1$. After 3 iterations, the optimal rank is identified within this interval.}
    \label{fig:searchspace}
\end{figure}

Throughout this procedure, the weights and rank coefficients are jointly optimized, ensuring that both the model parameters and the rank configuration are adapted to minimize the overall loss. The loss function can incorporate not only the reconstruction or decomposition error but also a regularization component that penalizes higher ranks, thereby promoting model compression.

At the conclusion of the search, the final rank configuration $\bar{\mathbf{r}} = (\bar{r}_1, \ldots, \bar{r}_n)$ is validated using a cross-entropy loss or another appropriate metric on a held-out validation set. This step ensures that the selected ranks yield not only a compact model but also satisfactory predictive performance. The balance between compression and accuracy can be tuned by adjusting the regularization parameters $\gamma$ and $\beta$ in the loss formulation.

This multi-step, progressive rank search method offers several advantages over traditional approaches. By systematically narrowing the search space and refining the sampling granularity, it combines the thoroughness of exhaustive search with the efficiency of adaptive optimization. The method is capable of escaping the limitations imposed by fixed candidate sets and can converge to globally optimal or near-optimal rank configurations. Figure~\ref{fig:searchspace} illustrates the evolution of the search space for a single layer: the process begins with a wide interval and large step size, then successively narrows and refines the search until the optimal rank is identified with maximal precision.

\begin{algorithm}[t!]
\caption{Rank adapt tENsor dEcomposition (\approach)}
\label{Algo2}
\textbf{Input:} Pretrained model $M$, Training data $X$, Rank lower bounds $\bs{Lb}=\{Lb_1,\ldots,Lb_n\}$ and upper bounds $\bs{Ub}=\{Ub_1,\ldots,Ub_n\}$, Number of iterations $T$, Step size $s > 1$, Factor $f$\;

\textbf{Initialize:} $\forall i, \mathcal{R}_i \gets \{r \mid r=Lb_i + ks, \text{ for } k\in\mathbb{N}, \text{ and } Lb_i\leq r\leq Ub_i\};$

\While{$s > 1$}{
    \For{$i \in \{1,\ldots,n\}$}{
      \For{$r \in \mathcal{R}_i$}{
          \For{$t = 1$ \textbf{to} $T$}{
            $\hat{\tensor{W}}^{(r)}_{i} \gets update(\hat{\tensor{W}}^{(r)}_{i})$\tcp*[r]{Eq. \eqref{eq:WU}}
            $\alpha^{(r)}_{i} \gets update(\alpha^{(r)}_{i})$\tcp*[r]{Eq. \eqref{eq:PU}}
          }
      }
    }

    $s \gets \left\lfloor \frac{s}{f} \right\rfloor$\;

    \For{$i \in \{1,\ldots,n\}$}{
      $\bar{r}_i \gets argmax_{r \in \mathcal{R}_i}(\text{softmax}(\alpha_i^{(r)}))$\;
      $Lb_i \gets \bar{r}_i - \frac{s}{2}$\;
      $Ub_i \gets \bar{r}_i + \frac{s}{2}$\;
      $\mathcal{R}_i \gets \{r \mid r=Lb_i + ks, \text{ for } k\in\mathbb{N}, \text{ and } Lb_i\leq r\leq Ub_i\}$\;
    }

}

\textbf{Output:} Decomposed model $M^*$  by minimizing $\mathcal{L}_f(\setw{W}^{\bs{\setr^\star}})$ using $X$ \tcp*[r]{Eq.~\eqref{eq:fine}}

\end{algorithm}
\subsection{Final Decomposition and Fine-Tuning}
The optimal ranks for decomposing the tensor weights for each layer, denoted as  $\bs{\setr^\star}=\{r^\star_1,\ldots,r^\star_n\}$, are determined from these final sets and used to construct the decomposed network. To ensure the decomposed model replicates the behavior of the original model, it is crucial that the layers not only align their decomposed weights with the original weights but also produce the same outputs. To achieve this, our fine-tuning loss ($\mathcal{L}_f$) consists of two components: a cross entropy loss ($\mathcal{L}_{ce}$) and distillation loss ($\mathcal{L}_{diss}$). The cross entropy loss adjusts the model's weights based on the training data labels. Distillation loss aligns the decomposed weights with the original weights by minimized Frobenius distance, and enforces consistency between the outputs of the original and decomposed layers. The distillation and fine-tuning losses are defined as follows:

\begin{equation}
\label{eq:fine}
\begin{aligned}
\mathcal{L}_{diss}(\setw{W}^{\bs{\setr^\star}}) = & \sum_{i=1}^{n} \left\|\tensor{W}_i - \hat{\tensor{W}}_{i}^{(r_i^\star)}\right\|^2_F  + \sum_{x\in X} \sum_{i=1}^{n} \left\|O_{i}(x) - D_{i}(x)\right\|^2_F,
\end{aligned}
\end{equation}

\begin{equation}
    \label{eq:fine}
\mathcal{L}_f(\setw{W}^{\bs{\setr^\star}}) = \mathcal{L}_{ce} + \lambda \mathcal{L}_{diss},
\end{equation}

where $X$ is the training set, $O_i(.)$ and $D_i(.)$ are the outputs of layer $i$ of the original model and the decomposed one, respectively and $\lambda$ is hyperparameter to control combination of losses. In this approach, the original model serves as the teacher model and the decomposed model acts as the student model. The pseudocode for the overall procedure retracing these steps is presented in Algorithm \ref{Algo2}.

\section{Experiments}

\subsection{Experimental Setup}
We evaluate \approach{}\footnote{The code  is available for research purposes at \url{https://github.com/aah94/RENE}} on 3 datasets including CIFAR-10/100 \cite{krizhevsky2009learning} and ImageNet-1K \cite{DengCHI14}. To prevent convergence collapse during the updating of Eq.\eqref{eq:TotalLoss} and Eq.\eqref{eq:TotalLossalpha}, we initially update only the weights for several iterations before jointly updating both weights and rank coefficients in an iterative manner. Each experiment is performed five times, and the best result from the fine-tuning step is reported. For TT decomposition, due to computational resource constraints, we assume that the two TT ranks are equal. In the search phase of \approach, for CIFAR-10/100, we set the initial rank space to $\{10,\ldots,100\}$ with a step size of $s=10$, which corresponds to 2 search steps. For ImageNet-1K, we set the initial rank space to $\{50,\ldots,850\}$ with the step size $s=100$, corresponding to 3 search steps. Across all datasets, we use $f=10$. We used the standard SGD optimizer with Nesterov momentum set to 0.9, and hyperparameters $\lambda$, $\gamma$ and $\beta$ set to 0.5, 0.4 and 0.8, respectively. The initial learning rates were 0.001 for CIFAR-10/100 and 0.0001 for ImageNet-1K. For the fine-tuning step we consider learning rate 0.00001 for all experiments and grid search with cross-validation is employed to select all hyperparameters, optimizing model performance based on validation accuracy. For comparing different approaches, the TOP-1 accuracy is used to compare the performance of the compressed model against the original uncompressed model. Additionally, we consider the gain in floating operations per second (FLOPs) and the compression rate.

\subsection{Experimental Results}
The following sections present a comprehensive analysis of \approach{}'s performance and compression capabilities across various models and datasets. 

\subsubsection{\textbf{Performance and Compression Analysis.}}

For the initial evaluation, we tested \approach{} on CIFAR-10 using the ResNet-20 and VGG-16 models, with the results presented in Table~\ref{tab:cifar10_results}. \approach{} with both CP and TT decomposition techniques yields competitive results compared to state-of-the-art methods. Using ResNet-20 as the original model, \approach{} with CP decomposition achieves 1.24\% and 1.52\% greater reduction of FLOPs and parameters, respectively, compared to the HALOC method \cite{xiao2023haloc}. Additionally, \approach{} with TT decomposition improves accuracy by 0.08\%  over the original uncompressed model. This suggests that our approach has effectively reduced the number of parameters of the original model, leading to a better generalization.  Furthermore, with the VGG-16 model, \approach{} achieves significant compression rates while preserving performance. For instance, using \approach{} with CP decomposition reduces FLOPs by 85.23\% and parameters by 98.6\%. Furthermore, applying \approach{} with TT decomposition on VGG-16 improves generalization, resulting in a  0.04\% increase in TOP-1 accuracy compared to the original uncompressed model.

The results on the ImageNet-1K dataset are presented in Table~\ref{tab:ImageNet-1K_results}, where we evaluated \approach{} using ResNet-18 and MobileNetV2 models. For ResNet-18, our approach with CP decomposition yields competitive results, while TT decomposition outperformed other methods, achieving state-of-the-art performance across all metrics, including Top-1 accuracy, reduction in FLOPs, and parameters.  With MobileNetV2, the CP method did not yield high performance, likely due to the model's reliance on depthwise convolution, which does not significantly benefit from decomposition in certain dimensions. However, \approach{} with TT decomposition demonstrated superior compression results, achieving 1.86\% and 2.31\% greater reductions in FLOPs and parameters, respectively, along with competitive Top-1 accuracy. Our results underscore the importance of selecting the appropriate decomposition method based on the model's complexity. Our experiments indicate that TT decomposition is more effective for compressing higher-complexity models, such as those trained on the ImageNet-1K dataset, while CP decomposition excels in compressing lower-complexity models, like those classically used on CIFAR-10. 

\begin{table}[t]
    \centering
    \begin{minipage}[t]{0.48\textwidth}  
        \centering
        \small
        \setlength{\tabcolsep}{4pt}
        \renewcommand{\arraystretch}{1.1}
        \caption{Results of different compression approaches for ResNet-20 and VGG-16 on CIFAR-10. C.T and A.R stand for \textit{compression technique} and \textit{automatic rank}, respectively.}
        \label{tab:cifar10_results}
        \vspace{2pt}\begin{adjustbox}{width=\textwidth}
        \begin{tabular}{l c c c c c}
        \toprule
        \textbf{Method} & \textbf{C.T} & \textbf{A.R} & \textbf{Top-1} & \textbf{FLOPs (↓\%)} & \textbf{Comp. Rate} \\
        \midrule
        ResNet-20 & Original & {-} & 91.25 & {-} & {-} \\
        \midrule
        \textbf{\approach(CP)} & Low-rank & \ding{51} & 90.82 & \textbf{73.44} & \textbf{77.62} \\
        \textbf{\approach(TT)} & Low-rank & \ding{51} & \textbf{91.40} & 70.4 & 72.28 \\
        \cdashline{1-6}
        HALOC \cite{xiao2023haloc} & Low-rank & \ding{51} & 91.32 & 72.20 & 76.10 \\
        ALDS \cite{liebenwein2021compressing} & Low-rank & \ding{51} & 90.92 & 67.86 & 74.91 \\
        LCNN \cite{idelbayev2020low} & Low-rank & \ding{51} & 90.13 & 66.78 & 65.38 \\
        PSTR-S \cite{li2021heuristic} & Low-rank & \ding{51} & 90.80 & 65.00 & 60.87 \\ 
        Std. Tucker \cite{kim2015compression} & Low-rank & \ding{55} & 87.41 & 62.00 & 61.54 \\ 
        \midrule
        VGG-16 & Original & {-} & 92.78 & {-} & {-} \\
        \midrule
        \textbf{\approach(CP)} & Low-rank & \ding{51} & 92.51 & 86.23 & \textbf{98.60} \\
        \textbf{\approach(TT)} & Low-rank & \ding{51} & \textbf{93.20} & 86.10 & 95.51 \\
        \cdashline{1-6}
        HALOC \cite{xiao2023haloc}& Low-rank & \ding{51} & 93.16 & \textbf{86.44} & 98.56 \\
        ALDS \cite{liebenwein2021compressing}& Low-rank & \ding{51} & 92.67 & 86.23 & 95.77 \\
        LCNN \cite{idelbayev2020low}& Low-rank & \ding{51} & 92.72 & 85.47 & 91.14 \\
        DECORE \cite{alwani2022decore} & Pruning & {-} & 92.44 & 81.50 & 96.60 \\
        Spike-Thrift \cite{kundu2021spike} & Pruning & {-} & 91.79 & 80.00 & 97.01 \\ 
        \bottomrule
        \end{tabular}
        \end{adjustbox}
    \end{minipage}
    \hfill
    \begin{minipage}[t]{0.48\textwidth}  
        \centering
        \small
        \setlength{\tabcolsep}{4pt}
        \renewcommand{\arraystretch}{1.1}
        \caption{Results of different compression approaches for ResNet-18 and MobileNetV2 on ImageNet-1K.}
        \label{tab:ImageNet-1K_results}
        \vspace{6pt}
        \begin{adjustbox}{width=\textwidth}
        \begin{tabular}{l c c c c c}
        \toprule
        \textbf{Method} & \textbf{C.T} & \textbf{A.R} & \textbf{Top-1} & \textbf{FLOPs (↓\%)} & \textbf{Comp. Rate} \\
        \midrule
        ResNet-18 & Original & {-} & 69.75 & {-} & {-} \\
        \midrule
        \textbf{\approach(CP)} & Low-rank & \ding{51} & 68.46 & 57.1 & 66.2 \\
        \textbf{\approach(TT)} & Low-rank & \ding{51} & \textbf{70.88} & \textbf{68.9} & \textbf{67.1} \\
        \cdashline{1-6}
        HALOC \cite{xiao2023haloc}& Low-rank & \ding{51} & 70.65 & 66.16 & 63.64 \\
        ALDS \cite{liebenwein2021compressing}& Low-rank & \ding{51} & 69.22 & 43.51 & 66.70 \\
        TETD \cite{yin2021towards} & Low-rank & \ding{55} & 69.00 & 59.51 & 60.00 \\ 
        Stable EPC \cite{phan2020stable} & Low-rank & \ding{51} & 68.50 & 59.51 & 61.00 \\ 
        MUSCO \cite{gusak2019automated} & Low-rank & \ding{55} & 69.29 & 58.67 & 60.50 \\ 
        CHEX \cite{hou2022chex} & Pruning & {-} & 69.60 & 43.38 & 59.00 \\ 
        EE \cite{zhang2021exploration} & Pruning & {-} & 68.27 & 46.60 & 58.00 \\ 
        SCOP \cite{tang2020scop} & Pruning & {-} & 69.18 & 38.80 & 39.30 \\
        \midrule
        MobileNetV2 & Original & {-} & 71.85 & {-} & {-} \\
        \midrule
        \textbf{\approach(CP)} & Low-rank & \ding{51} & 65.39 & 11.78 & \textbf{51.6} \\
        \textbf{\approach(TT)} & Low-rank & \ding{51} & 70.1 & \textbf{26.7} & 42.34 \\
        \cdashline{1-6}
        HALOC \cite{xiao2023haloc}& Low-rank & \ding{51} & \textbf{70.98} & 24.84 & 40.03 \\
        ALDS \cite{liebenwein2021compressing}& Low-rank & \ding{51} & 70.32 & 11.01 & 32.97 \\
        HOSA \cite{tang2020scop} & Pruning & {-} & 64.43 & 43.65 & 91.14 \\
        DCP \cite{chatzikonstantinou2020neural} & Pruning & {-} & 64.22 & 44.75 & 96.60 \\
        FT \cite{zhuang2018discrimination} & Pruning & {-} & 70.12 & 20.23 & 21.31 \\
        \bottomrule
        \end{tabular}
        \end{adjustbox}
    \end{minipage}
\end{table}

\subsubsection{\textbf{Automatic vs. Manual Rank Selection.}}

\begin{figure}[b!]
    \centering
    \begin{minipage}{0.48\textwidth}
        \centering
        \includegraphics[width=\textwidth]{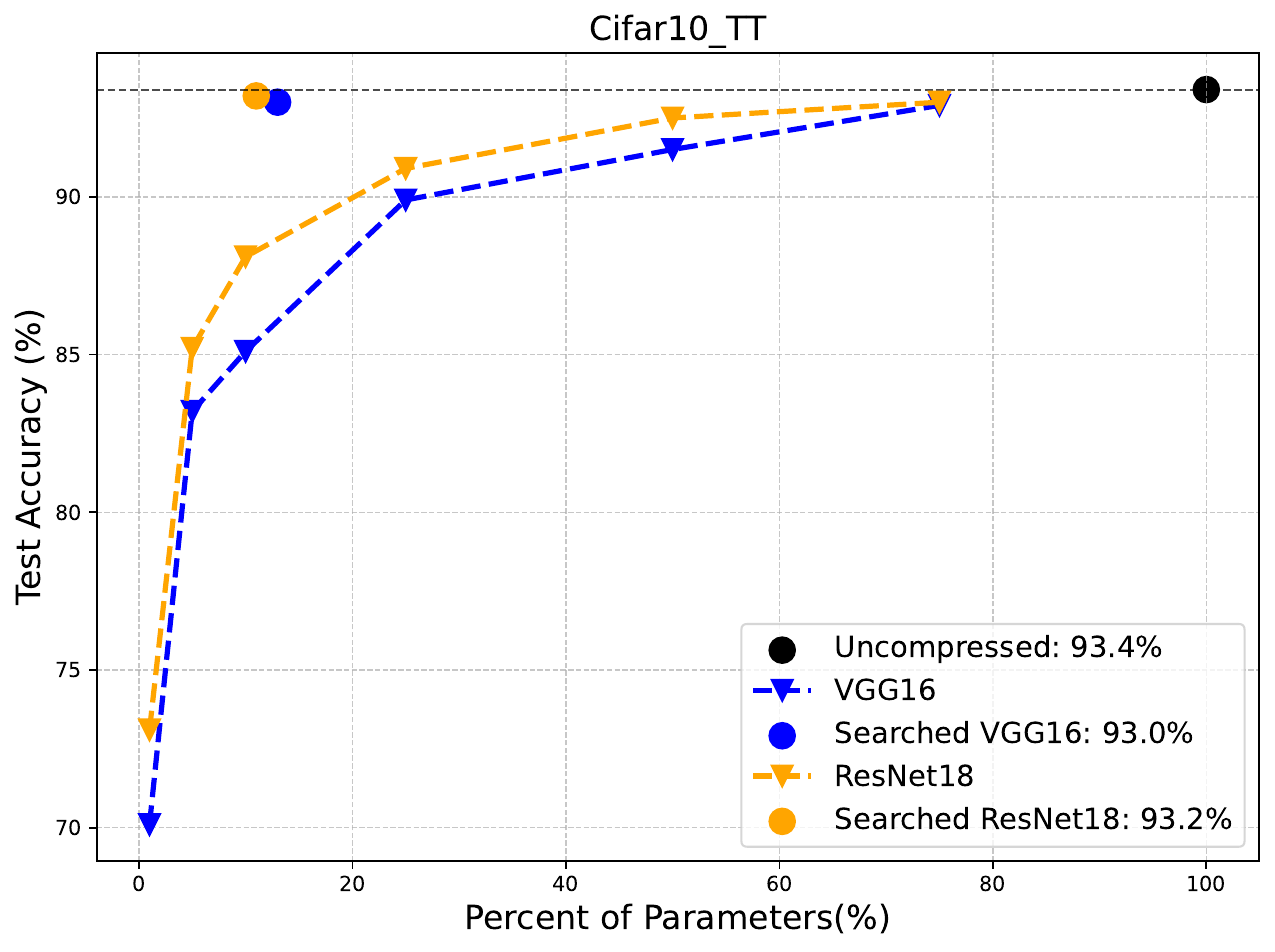}
    \end{minipage}
    \hfill
    \begin{minipage}{0.48\textwidth}
        \centering
        \includegraphics[width=\textwidth]{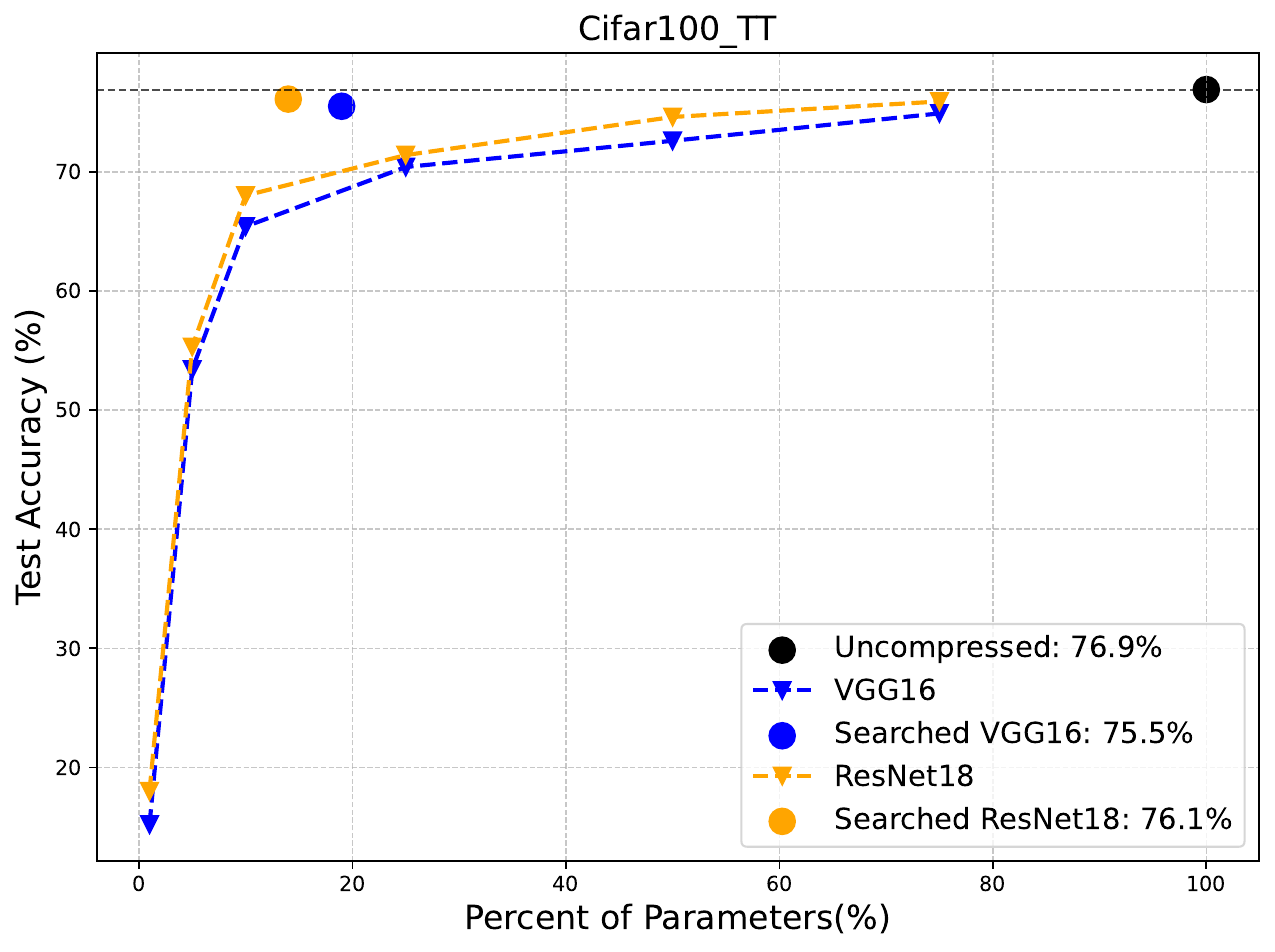}
    \end{minipage}
    \caption{Search vs Manual: Compression results for manual setting at different levels compression, compare to searched setting (left CIFAR10 and right CIFAR100).}
    \label{fig:comparison}
\end{figure}
We now examine the effectiveness of our rank search process compared to manual rank setting. In this experiment, we used pretrained ResNet18 and VGG16 models on the CIFAR-10 and CIFAR-100 datasets. For manual rank setting, we apply the TT decomposition and fix the rank across all layers to achieve a decomposed model with a specific percentage of the initial model's parameters, chosen from the set $\{1,5,10,25,50,75\}$. All models are pre-trained on the ImageNet-1K dataset and fine-tuned for 20 epochs. Figure \ref{fig:comparison} presents these results. As shown, increasing the number of ranks (or equivalently, increasing the percentage of parameters of the decomposed model) improves the performance of both the decomposed VGG16 and ResNet18 models. When the decomposed models have 75\% of the parameters of the initial models, the performance almost matches that of the original pretrained models. With \approach, We achieve comparable results while compressing the model by more than 80\% on both ResNet18 and VGG16 across both datasets. These results indicate that fixing the ranks across layers is suboptimal. In contrast, \approach{} enables the automatic selection of ranks across different layers, achieving a good compression rate without significant performance loss.

\subsubsection{\textbf{Rank selection}}

\begin{figure}[!]
           \centering
        \includegraphics[width=.6\textwidth]{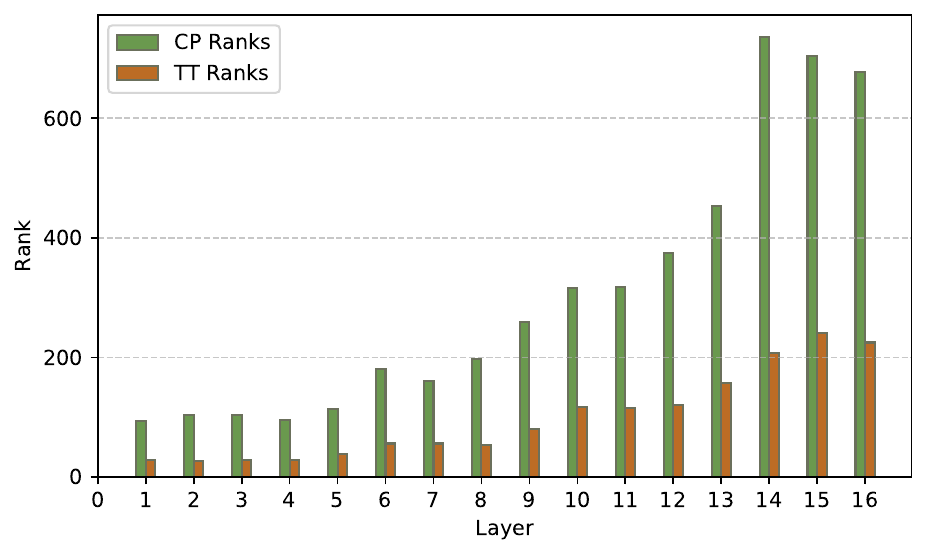}
        \captionof{figure}{Distribution of ranks achieved using CP and TT decompositions on ResNet-18 for the ImageNet-1K dataset.}
        \label{fig:ranks_distribution}
        \end{figure}

Figure \ref{fig:ranks_distribution} illustrates the selected ranks for both CP and TT decompositions using ResNet-18 as the original model on the ImageNet-1K dataset, highlighting that CP ranks are generally larger than those of TT.

 This difference arises from the inherent characteristics of the decomposition methods: CP decomposition tends to produce larger ranks because it decomposes the tensor into a sum of rank-one tensors, capturing more detailed interactions but potentially leading to higher complexity. In contrast, TT decomposition typically results in smaller ranks due to its chain-like structure, which can lead to more compact representations and potentially better compression. The distribution of ranks reveals that even among layers of the same dimensions, the effective ranks can differ. This reflects the varying contributions of each layer to the model's performance. Some layers may capture more complex features, requiring higher ranks, while others may focus on simpler features, allowing for lower ranks. These results are in line with the case of selecting the ranks manually and the same over all layers that were been presented in the previous section.

\subsubsection{\textbf{Double Compression}}

In this experiment, we investigate the effects of double compression by applying \approach{} in conjunction with knowledge distillation. Our goal is to assess whether combining these two compression techniques can yield further reductions in model size and computational requirements without sacrificing performance. We focus on TT decomposition for this analysis, using two datasets: CIFAR-100 and ImageNet-1K.

For the CIFAR-100 dataset, we employ ResNet-56 as the teacher model and ResNet-20 as the student model. Similarly, for the ImageNet-1K dataset, ResNet-34 serves as the teacher model, while ResNet-18 acts as the student model. The distillation process involves training the student model to mimic the behavior of the larger, more complex teacher model, thereby transferring knowledge and improving performance.
\begin{table}[!]
    \centering
        \captionof{table}{Double compression: \approach{} with distillation on CIFAR-100 and ImageNet-1K. The notations T and S denote the teacher and student, respectively.}
        \label{tab:distillation_compression}
        \begin{tabular}{l c c c}
            \toprule
            \multicolumn{4}{c}{\textbf{CIFAR-100 (T: ResNet56 (72.34\%), S: ResNet20 (69.6\%))}} \\
            \midrule
            \textbf{Method} & \textbf{Top-1 (\%)} & \textbf{FLOPs (\%)} & \textbf{Comp. rate (\%)} \\
            \midrule
            Distillation \cite{wang2023improving} & \textbf{72.53} & 67.7 & 68.24 \\
            \approach(Teacher) & 72.23 & 64.23 & 61.75 \\
            \approach(Student) & 72.46 & \textbf{89.01} & \textbf{86.54} \\
            \midrule
            \multicolumn{4}{c}{\textbf{ImageNet-1K (T: ResNet34 (73.31\%), S: ResNet18 (69.76\%))}} \\
            \midrule
            \textbf{Method} & \textbf{Top-1 (\%)} & \textbf{FLOPs (\%)} & \textbf{Params (\%)} \\
            \midrule
            Distillation \cite{wang2023improving} & 71.98 & 50.27 & 46.33 \\
            \approach(Teacher) & \textbf{73.23} & 59.91 & 63.46 \\
            \approach(Student) & 71.9 & \textbf{76.77} & \textbf{78.69} \\
            \bottomrule
        \end{tabular}
\end{table}

After applying distillation, we further compress both the teacher and the distilled student models using \approach{}. The results, presented in Table \ref{tab:distillation_compression}, demonstrate that our decomposition method achieves competitive performance compared to distillation alone for both the teacher and student models. Notably, when applying \approach{} to the distilled student model, we achieve a significant reduction in both parameters and computational complexity. Specifically, on the ImageNet-1K dataset, the decomposed distilled student model reduces parameters by 78.69\% and FLOPs by 76.77\% compared to the original teacher model.

This double compression approach not only maintains the accuracy of the original model but also highlights the potential for substantial reductions in model size and computational requirements. These findings underscore the effectiveness of combining distillation with decomposition techniques to achieve efficient and high-performing compressed models.

\section{Conclusion}

In this paper, we presented an approach for compressing deep neural networks through decomposition and optimal rank selection. Our solution stands out with two key features: it considers all layers during the optimization process, aiming for high compression rates without compromising accuracy by identifying the optimal rank pattern across layers. This approach capitalizes on the varying contributions of different layers to the model's inference, allowing for smaller ranks in less critical layers and determining the most effective rank pattern for each. To achieve significant compression, we explore a broad range of ranks, addressing the substantial memory challenges of this extensive exploration with a multistage rank search strategy. This strategy enables comprehensive exploration while ensuring efficient memory usage.  Our experimental results demonstrate that this approach effectively reduces the number of parameters and computational complexity, leading to better generalization and competitive performance across various models and datasets.

\bibliographystyle{splncs04}
\bibliography{Biblio}

\end{document}